\documentclass[10pt,twocolumn,letterpaper]{article}

\usepackage{cvpr}              

\usepackage{graphicx}
\usepackage{amsmath}
\usepackage{amssymb}
\usepackage{booktabs}
\usepackage{times}
\usepackage{microtype}
\usepackage{epsfig}
\usepackage[table,xcdraw]{xcolor}
\usepackage{caption}
\usepackage{xcolor}
\usepackage{float}
\usepackage{placeins}
\usepackage{color, colortbl}
\usepackage{stfloats}
\usepackage{enumitem}
\usepackage{tabularx}
\usepackage{xstring}
\usepackage{multirow}
\usepackage{xspace}
\usepackage{url}
\usepackage{subcaption}
\usepackage{arydshln}
\usepackage{xcolor}
\usepackage{bbm}
\usepackage[hang,flushmargin]{footmisc}
\usepackage{pifont}
\newcommand{\cmark}{\ding{51}}%
\usepackage[accsupp]{axessibility}

\newcommand{\et}[2]{${#1}^{\pm{#2}}$}
\newcommand{\etb}[2]{$\mathbf{{#1}}^{\pm{#2}}$}
\newcommand{\etr}[2]{$\textcolor{red}{{#1}}^{\pm{#2}}$}
\newcommand{\etbb}[2]{$\textcolor{blue}{{#1}}^{\pm{#2}}$}

\usepackage[pagebackref,breaklinks,colorlinks]{hyperref}

\usepackage[capitalize]{cleveref}
\crefname{section}{Sec.}{Secs.}
\Crefname{section}{Section}{Sections}
\Crefname{table}{Table}{Tables}
\crefname{table}{Tab.}{Tabs.}

\def\cvprPaperID{**}
\def\confName{****}
\def\confYear{****}

\begin{document}

\title{T2M-GPT: Generating Human Motion from Textual Descriptions with \\ Discrete Representations}

\author{Jianrong Zhang$^{1,3*}$, 
Yangsong Zhang$^{2,3*}$, 
Xiaodong Cun$^3$, Shaoli Huang$^3$, Yong Zhang$^3$ \\ Hongwei Zhao$^1$, Hongtao Lu$^2$, Xi Shen$^{3,\dagger}$ \\
\small $^{*}$Equal contribution \qquad $^{\dagger}$Corresponding author \\[2mm]
$^1$Jilin University \qquad
$^2$Shanghai Jiao Tong University \qquad $^3$Tencent AI Lab\\
}
\maketitle

\begin{abstract}
In this work, we investigate a simple and must-known conditional generative framework based on Vector Quantised-Variational AutoEncoder (VQ-VAE) and Generative Pre-trained Transformer (GPT) for human motion generation from textural descriptions. We show that a simple CNN-based VQ-VAE with commonly used training recipes (EMA and Code Reset) allows us to obtain high-quality discrete representations. For GPT, we incorporate a simple corruption strategy during the training to alleviate training-testing discrepancy. Despite its simplicity, our T2M-GPT shows better performance than competitive approaches, including recent diffusion-based approaches. For example, on HumanML3D, which is currently the largest dataset, we achieve comparable performance on the consistency between text and generated motion (R-Precision), but with FID 0.116 largely outperforming MotionDiffuse of 0.630. Additionally, we conduct analyses on HumanML3D and observe that the dataset size is a limitation of our approach. Our work suggests that VQ-VAE still remains a competitive approach for human motion generation. Our implementation is available on the project page: \url{https://mael-zys.github.io/T2M-GPT/}.
\end{abstract}

\section{Introduction}
\label{sec:intro}
\begin{figure}[tp]
    \centering
    \includegraphics[width=0.47\textwidth]{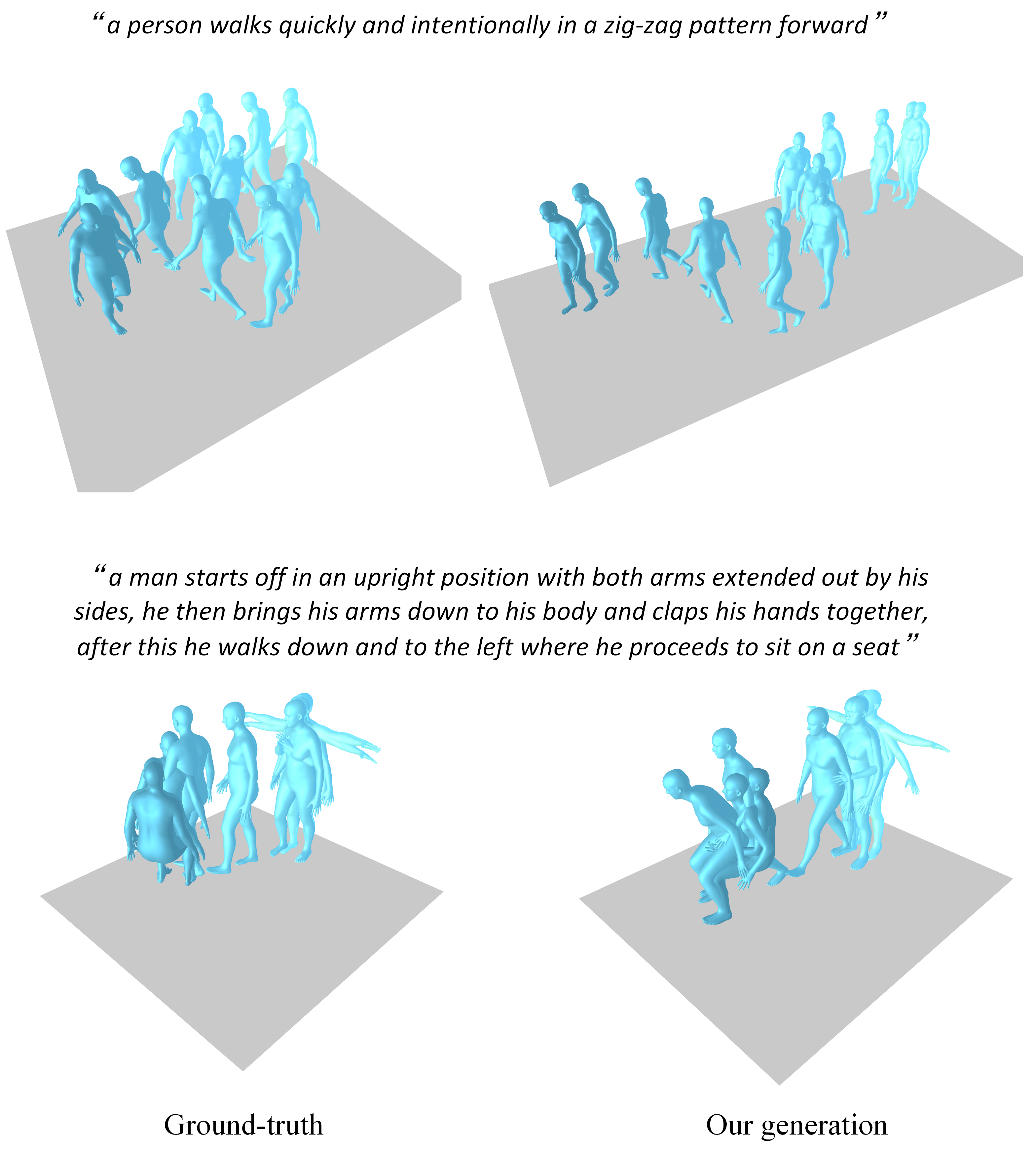}
    \caption{\textbf{Visual results on HumanML3D~\cite{guo2022generating}.} Our approach is able to generate precise and high-quality human motion consistent with challenging text descriptions. More visual results are on the \href{https://mael-zys.github.io/T2M-GPT/}{project page}.}
    \label{fig:teaser}
\end{figure}

Generating motion from textual descriptions can be used in numerous applications in the game industry, film-making, and animating robots. For example, a typical way to access new motion in the game industry is to perform motion capture, which is expensive. Therefore automatically generating motion from textual descriptions, which allows producing meaningful motion data, could save time and be more economical. 

Motion generation conditioned on natural language is challenging, as motion and text are from different modalities. The model is expected to learn precise mapping from the language space to the motion space. To this end, many works propose to learn a joint embedding for language and motion using auto-encoders~\cite{ahuja2019language2pose, ghosh2021synthesis, tevet2022motionclip} and VAEs~\cite{petrovich21actor,petrovich22temos}. MotionClip~\cite{tevet2022motionclip} aligns the motion space to CLIP~\cite{radford2021learning} space. ACTOR~\cite{petrovich21actor} and TEMOES~\cite{petrovich22temos} propose transformer-based VAEs for action-to-motion and text-to-motion respectively. These works show promising performances with simple descriptions and are limited to producing high-quality motion when textual descriptions become long and complicated. Guo \textit{et al.}~\cite{guo2022generating} and TM2T~\cite{chuan2022tm2t} aim to generate motion sequences with more challenging textual descriptions. However, both approaches are not straightforward, involve three stages for text-to-motion generation, and sometimes fail to generate high-quality motion consistent with the text (See Figure~\ref{fig:visual} and more visual results on the \href{https://mael-zys.github.io/T2M-GPT/}{project page}). 
Recently, diffusion-based models~\cite{ho2020denoising} have shown impressive results on image generation~\cite{rombach2022high}, which are then introduced to motion generation by MDM~\cite{tevet2022MDM} and MotionDiffuse~\cite{zhang2022motiondiffuse} and dominates text-to-motion generation task. However, we find that compared to classic approaches, such as VQ-VAE~\cite{van2017neural}, the performance gain of the diffusion-based approaches~\cite{zhang2022motiondiffuse,tevet2022MDM} might not be that significant. 
In this work, we are inspired by recent advances from learning the discrete representation for generation~\cite{van2017neural,williams2020hierarchical,esser2021taming,ramesh2021zero,ao2022rhythmic,dieleman2018challenge,dhariwal2020jukebox,posegpt} and investigate a simple and classic framework based on Vector Quantized Variational Autoencoders (VQ-VAE)~\cite{van2017neural} and Generative Pre-trained Transformer (GPT)~\cite{vaswani2017attention,radford2018improving} for text-to-motion generation.

Precisely, we propose a two-stage method for motion generation from textual descriptions. In stage 1, we use a standard 1D convolutional network to map motion sequences to discrete code indices. In stage 2, a standard GPT-like model~\cite{vaswani2017attention,radford2018improving} is learned to generate sequences of code indices from pre-trained text embedding.
We find that the naive training of VQ-VAE~\cite{van2017neural} suffers from code collapse. One effective solution is to leverage two standard recipes during the training: \textit{EMA} and \textit{Code Reset}. We provide a full analysis of different quantization strategies. For GPT, the next token prediction brings inconsistency between the training and inference. We observe that simply corrupting sequences during the training alleviates this discrepancy. Moreover, throughout the evolution of image generation, the size of the dataset has played an important role. We further explore the impact of dataset size on the performance of our model. The empirical analysis suggests that the performance of our model can potentially be improved with larger datasets.

Despite its simplicity, our approach can generate high-quality motion sequences that are consistent with challenging text descriptions (Figure~\ref{fig:teaser} and more on the \href{https://mael-zys.github.io/T2M-GPT/}{project page}). Empirically, we achieve comparable or even better performances than concurrent diffusion-based approaches MDM~\cite{tevet2022MDM} and HumanDiffuse~\cite{zhang2022motiondiffuse} on two widely used datasets: HumanML3D~\cite{guo2022generating} and KIT-ML~\cite{plappert2016kit}. For example, on HumanML3D, which is currently the largest dataset, we achieve comparable performance on the consistency between text and generated motion (R-Precision), but with FID 0.116 largely outperforming MotionDiffuse of 0.630.
We conduct comprehensive experiments to explore this area, and hope that these experiments and conclusions will contribute to future developments.

In summary, our contributions include: 
\begin{itemize}
    \item We present a simple yet effective approach for motion generation from textual descriptions. Our approach achieves state-of-the-art performance on HumanML3D~\cite{guo2022generating} and KIT-ML~\cite{plappert2016kit} datasets.
    \item We show that GPT-like models incorporating discrete representations still remain a very competitive approach for motion generation.
    \item We provide a detailed analysis of the impact of quantization strategies and dataset size. We show that a larger dataset might still offer a promising prospect to the community.
\end{itemize}

Our implementation is available on the \href{https://mael-zys.github.io/T2M-GPT/}{project page}.

\section{Related work}
\label{sec:related_work}

\paragraph{VQ-VAE.} Vector Quantized Variational Autoencoders (VQ-VAE), which is a variant of VAE~\cite{kingma2013auto}, is initially proposed in~\cite{van2017neural}. VQ-VAE is composed of an AutoEncoder architecture, which aims at learning reconstruction with discrete representations.  
Recently, VQ-VAE achieves promising performance on generative tasks across different modalities, which includes: image synthesis~\cite{williams2020hierarchical,esser2021taming}, text-to-image generation~\cite{ramesh2021zero}, speech gesture generation~\cite{ao2022rhythmic}, music generation~\cite{dieleman2018challenge,dhariwal2020jukebox} etc. The success of VQ-VAE for generation might be attributed to its decoupling of learning the discrete representation and the prior. 
A naive training of VQ-VAE suffers from the codebook collapse, \textit{i.e.}, only a number of codes are activated, which importantly limited the performances of the reconstruction as well as generation. To alleviate the problem, a number of techniques can be used during training, including stop-gradient along with some losses~\cite{van2017neural} to optimize the codebook, exponential moving average (EMA) for codebook update~\cite{williams2020hierarchical}, reset inactivated codes during the training (Code Reset~\cite{williams2020hierarchical}), etc. 

\paragraph{Human motion synthesis.} Research on human motion synthesis has a long history~\cite{badler1993simulating}. One of the most active research fields is human motion prediction, which aims at predicting the future motion sequence based on past observed motion. Approaches mainly focus on efficiently and effectively fusing spatial and temporal information to generate deterministic future motion through different models: RNN~\cite{fragkiadaki2015recurrent,martinez2017human,butepage2017deep,pavllo2018quaternet}, GAN~\cite{hernandez2019human,barsoum2018hp}, GCN~\cite{mao2019learning}, Attention~\cite{mao2020history}
 or even simply MLP~\cite{guo2022back,bouazizi2022motionmixer,du2023avatars}. Some approaches aim at generating diverse motion through VAE~\cite{habibie2017recurrent,yan2018mt,aliakbarian2020stochastic}. In addition to synthesizing motion conditioning on past motion, another related topic is generating motion “in-betweening” that takes both past and future poses and fills motion between them~\cite{harvey2018recurrent,kaufmann2020convolutional,harvey2020robust,duan2021single,tang2022real}.~\cite{pavllo2018quaternet} considers the generation of locomotion sequences from a given trajectory for simple actions such as: walking and running. Motion can also be generated with music to produce 3D dance motion~\cite{lee2019dancing,li2020learning,li2021ai,aristidou2021rhythm,chen2021choreomaster,siyao2022bailando}. For unconstrained generations,  ~\cite{yan2019convolutional} generates a long sequence altogether by transforming from a sequence of latent vectors sampled from a Gaussian process. In graphics literature, many works focus on animator control. Holden \textit{et al.}~\cite{holden2016deep} learn a convolutional autoencoder to reconstruct motion, the learned latent representation can be used to synthesize and edit motion.~\cite{holden2017phase} proposes phase functioned neural network to perform the control task.~\cite{starke2019neural} uses a deep auto-regressive framework to scene interaction behaviors. Starke \textit{et al.}~\cite{starke2022deepphase} proposes to reconstruct motion through periodic features, the learned periodic embedding improves motion synthesis. Recently, inspired by SinGAN~\cite{shaham2019singan} for image synthesis, Li \textit{et al.}~\cite{li2022ganimator} propose a generative model approach for motion synthesis from a single sequence.

\begin{figure*}[t]
  \begin{subfigure}[b]{1.3\columnwidth}
    \includegraphics[width=\linewidth]{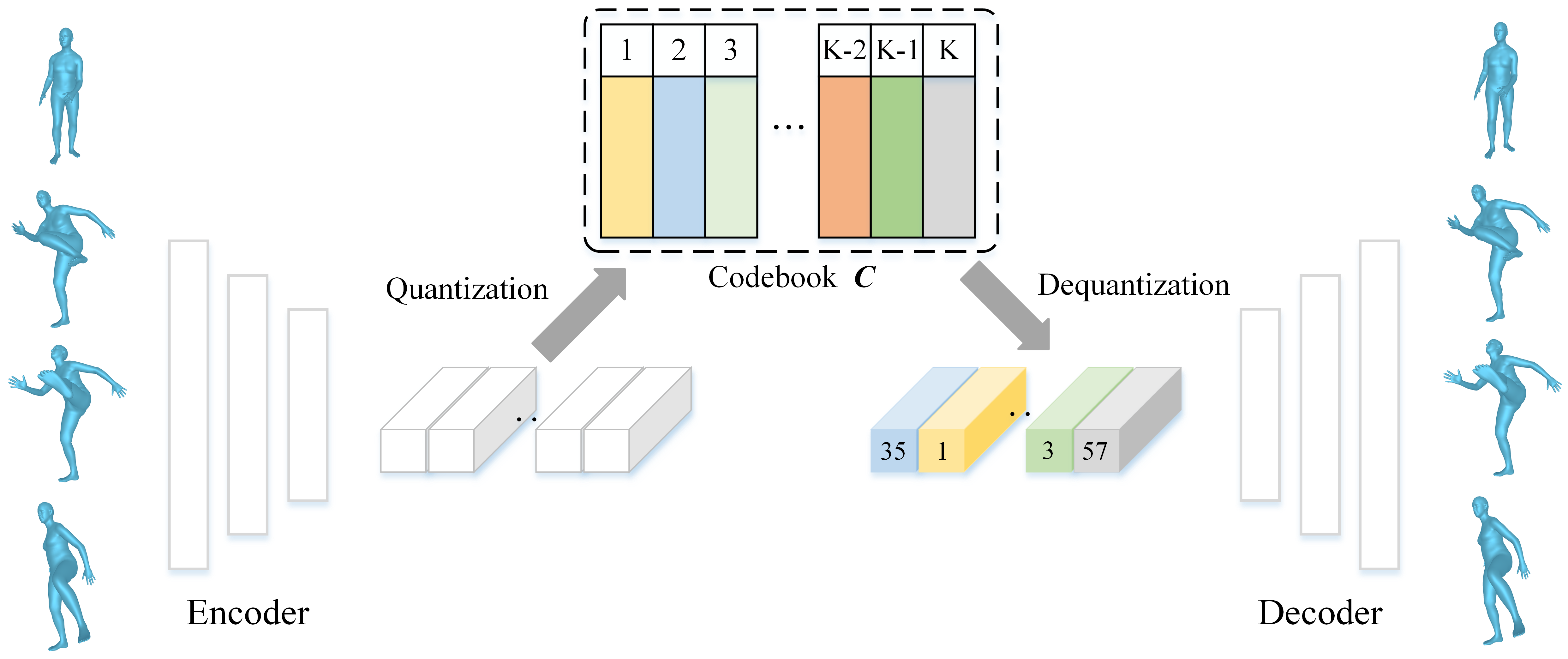}
    \caption{Motion VQ-VAE}
    \label{fig:vqvae}
  \end{subfigure}
  \hfill 
  \begin{subfigure}[b]{0.77\columnwidth}
    \includegraphics[width=\linewidth]{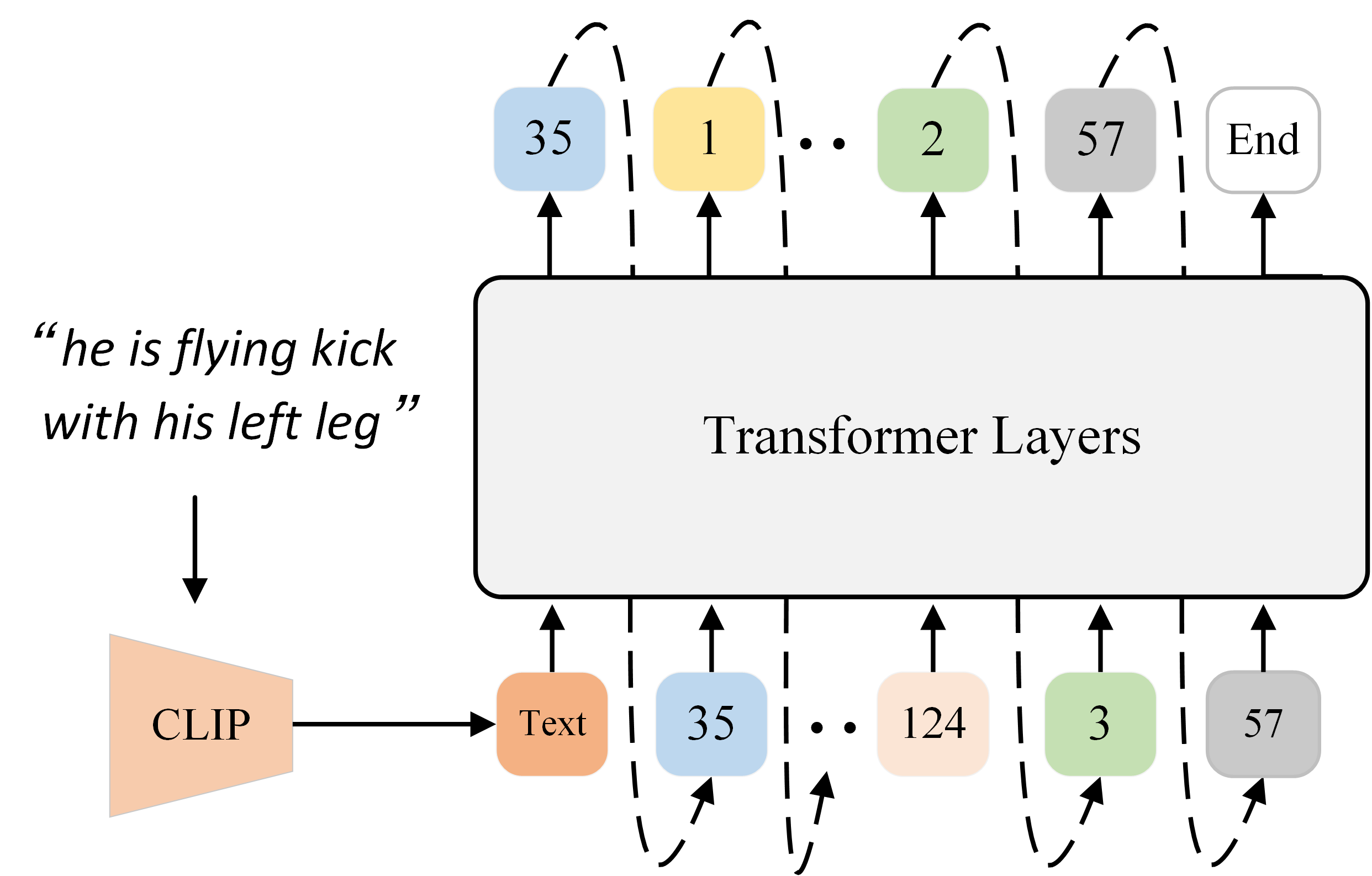}
    \caption{T2M-GPT}
    \label{fig:transformer}
  \end{subfigure}
  \caption{\textbf{Overview of our framework for text-driven motion generation.} It includes two modules: Motion VQ-VAE (Figure~\ref{fig:vqvae}) and T2M-GPT (Figure~\ref{fig:transformer}). In T2M-GPT, an additional learnable $\mathit{End}$ token is inserted to indicate the stop of the generation. During the inference, we first generate code indexes in an auto-regressive fashion and then obtain the motion using the decoder in Motion VQ-VAE.}
  \label{fig:overview}
\end{figure*}

\paragraph{Text-driven human motion generation.} Text-driven human motion generation aims at generating 3D human motion from textual descriptions. Text2Action~\cite{ahn2018text2action} trains
an RNN-based model to generate motion conditioned on a short text. Language2Pose~\cite{ahuja2019language2pose} employs a curriculum learning approach to learn a joint embedding space for both text and pose. The decoder can thus take text embedding to generate motion sequences. Ghost \textit{et al.}~\cite{ghosh2021synthesis} learn two manifold representations for the upper body and the lower body movements, which shows improved performance compared to Language2Pose~\cite{ahuja2019language2pose}. Similarly, MotionCLIP~\cite{tevet2022motionclip} also tends to align text and motion embedding but proposes to utilize CLIP~\cite{radford2021learning} as the text encoder and employ rendered images as extra supervision. It shows the ability to generate out-of-distribution motion and enable latent code editing. However, the generated motion sequences are not in high-quality and are without global translation. ACTOR~\cite{petrovich21actor} proposes a transformer-based VAE to generate motion in a non-autoregressive fashion from a pre-defined action class. TEMOS~\cite{petrovich22temos} extends the architecture of ACTOR~\cite{petrovich21actor} by introducing an additional text encoder and producing diverse motion sequences given text descriptions. TEMOS demonstrates its effect on KIT Motion-Language~\cite{plappert2016kit} with mainly short sentences and suffers from out-of-distribution descriptions~\cite{petrovich22temos}. TEACH~\cite{TEACH:3DV:2022} further extends TEMOS to generate temporal motion compositions from a series of natural language descriptions. Recently, a large-scale dataset HumanML3D is proposed in~\cite{guo2022generating}. Guo \textit{et al.}~\cite{guo2022generating} also propose to incorporate motion length prediction from text to produce motion with reasonable length. TM2T~\cite{chuan2022tm2t} considers text-to-motion and motion-to-text tasks. It also shows additional improvement can be obtained through jointly training both tasks. As concurrent works, diffusion-based models are introduced for text-to-motion generation by MDM~\cite{tevet2022MDM} and MotionDiffuse~\cite{zhang2022motiondiffuse}. In this work, we show that without any sophisticated designs, the classic VQ-VAE framework could achieve competitive or even better performance with a classical framework and some standard training recipes. 

\section{Method}
\label{sec:method}
Our goal is to generate high-quality motion that is consistent with text descriptions. The overall framework consists of two modules: Motion VQ-VAE and T2M-GPT, which is illustrated in Figure~\ref{fig:overview}. The former learns a mapping between motion data and discrete code sequences, the latter generates code indices conditioned on the text description. With the decoder in Motion VQ-VAE, we are able to recover the motion from the code indices. In Section~\ref{sec:vqvae}, we present the VQ-VAE module. The T2M-GPT is introduced in Section~\ref{sec:transformer}.

\subsection{Motion VQ-VAE}
\label{sec:vqvae}

VQ-VAE, proposed in~\cite{van2017neural}, enables the model to learn discrete representations for generative models. Given a motion sequence $X = [x_1, x_2, \ldots, x_T ]$ with $x_t \in \mathbb{R}^{d}$, where $T$ is the number of frames and $d$ is the dimension of the motion, we aim to recover the motion sequence through an autoencoder and a learnable codebook containing $K$ codes $C=\{c_k\}_{k=1}^K$ with $c_k \in \mathbb{R}^{d_c}$, where $d_c$ is the dimension of codes. The overview of VQ-VAE is presented in Figure~\ref{fig:vqvae}. With encoder and decoder of the autoencoder denoted by $E$ and $D$, the latent feature $Z$ can be computed as $Z = E(X)$ with $Z = [z_1, z_2, ..., z_{T/l}]$ and $z_i \in \mathbb{R}^{d_c}$, where $l$ represents the temporal downsampling rate of the encoder $E$. For $i$-th latent feature $z_i$, the quantization through $C$ is to find the most similar element in $C$, which can be properly written as:
\begin{equation}
\hat{z_i} =  \underset{c_k\in C}\arg\min\|z_i - c_k\|_2
\label{formula:1}
\end{equation}

\paragraph{Optimization goal.} To optimize VQ-VAE, the standard optimization goal~\cite{van2017neural} $\mathcal{L}_{\text{vq}}$ contains three components: a reconstruction loss $\mathcal{L}_{\text{re}}$, the embedding loss $\mathcal{L}_{embed}$ and the commitment loss $\mathcal{L}_{commit}$. 
\begin{equation}
\mathcal{L}_{vq} = \mathcal{L}_{re} + \underbrace{||\mathit{sg}[Z] - \hat{Z}||_2}_{\mathcal{L}_{embed}} + \beta \underbrace{||Z - \mathit{sg}[\hat{Z}]||_2}_{\mathcal{L}_{commit}}
\label{formula:2}
\end{equation}
where $\beta$ is a hyper-parameter for the commitment loss and $\mathit{sg}$ is the stop-gradient operator. For the reconstruction, we find that L1 smooth loss $\mathcal{L}_1^{smooth}$ performs best and an additional regularization on the velocity enhances the generation quality. Let $X_{\text{re}}$ be the reconstructed motion of $X$, i.e.,~$X_{\text{re}}=D(\hat{Z})$, V(X) be the velocity of $X$ where $V = [v_1, v_2, \ldots, v_{T-1}]$ with $v_i = x_{i+1} - x_i$. Therefore, the objective of the reconstruction is as follows:
\begin{equation}
\mathcal{L}_{re} = \mathcal{L}_1^{smooth}(X, X_{re}) + \alpha \mathcal{L}_1^{smooth}(V(X), V(X_{re}))
\label{formula:3}
\end{equation}
where $\alpha$ is a hyper-parameter to balance the two losses. We provide an ablation study on $\alpha$ as well as different reconstruction losses ($\mathcal{L}_1$, $\mathcal{L}_1^{smooth}$ and $\mathcal{L}_2$) in Section~\ref{sec:supp_vq_loss} of the appendix. 

\begin{figure}[t]
    \centering
    \includegraphics[scale=0.55]{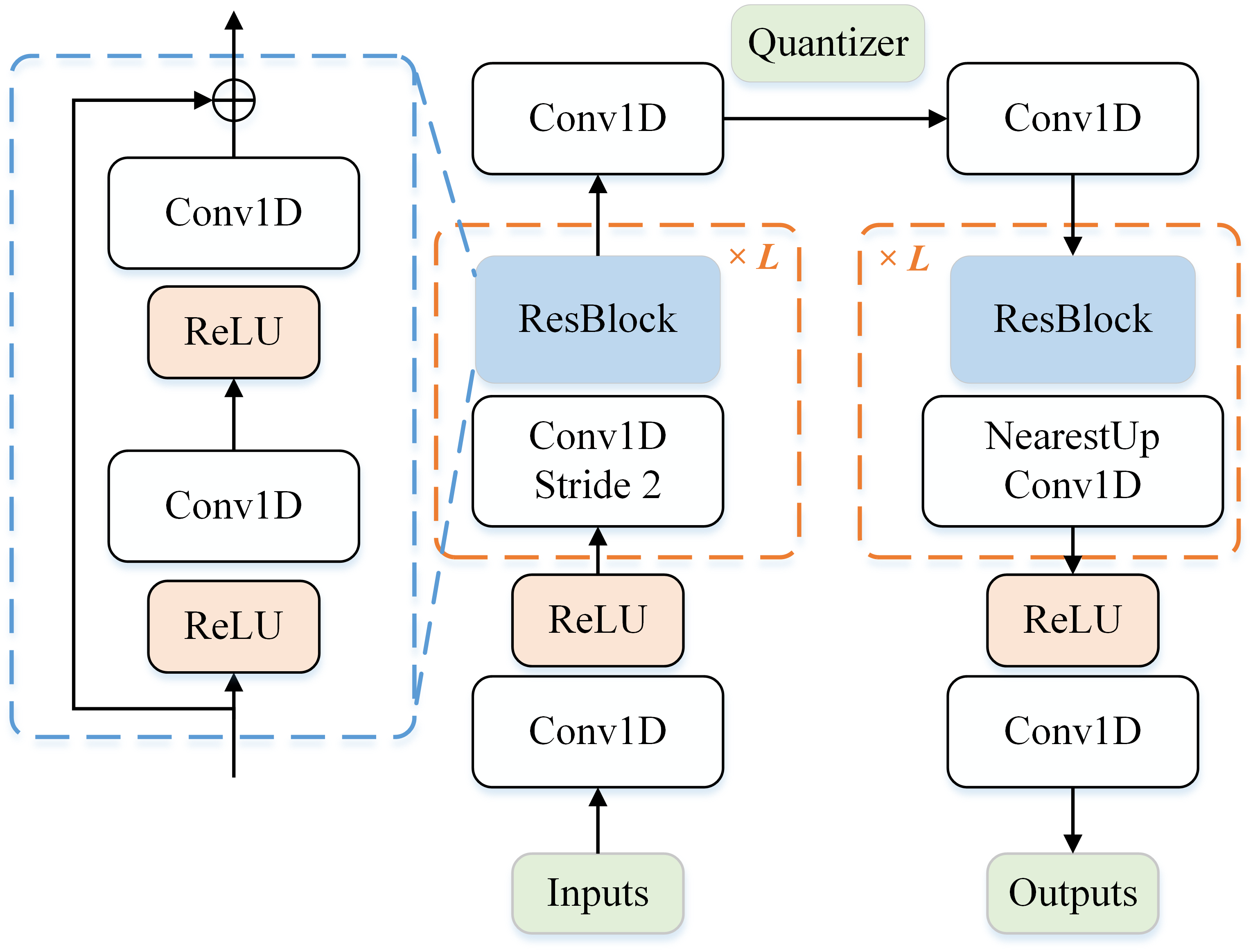}
    \caption{\textbf{Architecture of the motion VQ-VAE.} We use a standard CNN-based architecture with 1D convolution (\textit{Conv1D}), residual block (\textit{ResBlock}) and ReLU activation. `$L$' denotes the number of residual blocks. We use convolution with stride 2 and nearest interpolation for temporal downsampling and upsampling.}
    \label{fig:vqarch}
\end{figure}

\paragraph{Quantization strategy.} A naive training of VQ-VAE suffers from codebook collapse~\cite{van2017neural,razavi2019generating}. Two training recipes~\cite{razavi2019generating} are commonly used to improve the codebook utilization: exponential moving average \textit{(EMA)} and codebook reset \textit{(Code Reset)}. \textit{EMA} makes the codebook $C$ evolve smoothly: $C^t \leftarrow \lambda C^{t-1} + (1 - \lambda) C^{t}$, where $C^t$ is the codebook at iteration $t$ and $\lambda$ is the exponential moving constant. \textit{Code Reset} finds inactivate codes during the training and reassigns them according to input data. We provide an ablation study on the quantization strategy in Section~\ref{sec:discuss}.

\paragraph{Architecture.} We use a simple convolutional architecture composed of 1D convolution, residual block~\cite{he2016deep}, and ReLU. Our VQ-VAE architecture is illustrated in Figure~\ref{fig:vqarch}. The architecture is inspired by~\cite{esser2021taming,siyao2022bailando}. We use convolution with stride 2 and nearest interpolation for temporal downsampling and upsampling respectively. The downsampling rate is thus $l = 2^L$, where $L$ denotes the number of residual blocks. We provide an ablation study on the architecture in Section~\ref{sec:discuss}. The detail of the architecture is provided in Section~\ref{sec:supp_vq_arch} of the appendix.

\subsection{T2M-GPT}
\label{sec:transformer}
With a learned motion VQ-VAE, a motion sequence $X = [x_1, x_2, \ldots, x_T ]$ can be mapped to a sequence of indices $S = [s_1, s_2, \ldots, s_{T/l}, \mathit{End}]$ with $s_i \in [1, 2, \ldots, s_{T/l}]$, which are indices from the learned codebook. Note that a special $\mathit{End}$ token is added to indicate the stop of the motion, which is different from~\cite{guo2022generating} that leverages an extra module to predict motion length. By projecting $S$ back to their corresponding codebook entries, we obtain $\hat{Z} = [\hat{z}_1, \hat{z}_2, \ldots, \hat{z}_{T/l}]$ with $\hat{z}_i = c_{s_i}$, which can be decoded to a motion $X_{re}$ through the decoder $D$. Therefore, text-to-motion generation can be formulated as an autoregressive next-index prediction: given previous $i-1$ indices, i.e.,~$S_{< i}$, and text condition $c$, we aim to predict the distribution of possible next indices $p(S_i|c,S_{< i})$, which can be addressed with transformer~\cite{vaswani2017attention}. The overview of our transformer is shown in Figure~\ref{fig:transformer}.

\paragraph{Optimization goal.} Denoting the likelihood of the full sequence as $p(S|c) = \prod^{|S|}_{i=1}p(S_i|c,S_{< i})$, we directly
maximize the log-likelihood of the data distribution:
\begin{equation}
	\mathcal{L}_{trans} = \mathbb{E}_{S\sim p(S)}[-\log p(S|c)]
	\label{formula:5}
\end{equation}

We leverage CLIP~\cite{radford2021learning} to extract text embedding $c$, which has shown its effectiveness in relevant tasks~\cite{tevet2022motionclip,ramesh2022hierarchical,crowson2022vqgan}. 

\paragraph{Causal Self-attention.} We apply the causal self-attention~\cite{radford2018improving} in T2M-GPT. Precisely, the output of the causal self-attention is calculated as follows:
\begin{equation}
	\text{Attention} = \text{Softmax} \left( \frac{QK^T \times mask}{\sqrt{d_k}} \right)
	\label{formula:causal}
\end{equation}
where $Q \in \mathbb{R}^{T \times d_k}$ and $K \in \mathbb{R}^{T \times d_k}$ are query and key respectively, while $mask$ is the causal mask with $mask_{i,j} = -\infty \times \mathbf{1}\left( i < j\right) + \mathbf{1}\left( i \geq j \right)$, where $\mathbf{1}(\cdot)$ is the indicator function.
This causal mask ensures that future information is not allowed to attend the calculation of current tokens. For inference, we start from the text embedding and generate indices in an autoregressive fashion, the generation process will be stopped if the model predicts the $\mathit{End}$ token. Note that we are able to generate diverse motions by sampling from the predicted distributions given by the transformer.

\paragraph{Corrupted sequences for the training-testing discrepancy.} There is a discrepancy between training and testing. For training, $i-1$ correct indices are used to predict the next index. While for inference, there is no guarantee that indices serving as conditions are correct. To address this problem, we adopt a simple data augmentation strategy:  we replace $\tau\times100\%$ ground-truth code indices with random ones during training. $\tau$ can be a hyper-parameter or randomly sampled from $\tau \in \mathcal{U}[0, 1]$. We provide an ablation study on this strategy in Section~\ref{sec:supp_rep_ratio} of the appendix.  

\begin{table*}[t]
    \centering
    \scalebox{0.97}{

    \begin{tabular}{l c c c c c c c}
    \toprule
    \multirow{2}{*}{Methods}  & \multicolumn{3}{c}{R-Precision $\uparrow$} & \multirow{2}{*}{FID $\downarrow$} & \multirow{2}{*}{MM-Dist $\downarrow$} & \multirow{2}{*}{Diversity $\uparrow$} & \multirow{2}{*}{MModality $\uparrow$}\\

    \cline{2-4}
    ~ & Top-1 & Top-2 & Top-3 \\
    
    \midrule

        \textbf{Real motion} & \et{0.511}{.003} & \et{0.703}{.003} & \et{0.797}{.002} & \et{0.002}{.000} & \et{2.974}{.008} & \et{9.503}{.065} & -  \\
        Our VQ-VAE \small{(Recons.)} & \et{0.501}{.002} & \et{0.692}{.002} & \et{0.785}{.002} & \et{0.070}{.001}& \et{3.072}{.009} & \et{9.593}{.079} & -  \\ 
    \midrule
        Seq2Seq~\cite{lin2018generating} & \et{0.180}{.002} & \et{0.300}{.002} & \et{0.396}{.002} & \et{11.75}{.035} & \et{5.529}{.007} & \et{6.223}{.061}  & -  \\

        Language2Pose~\cite{ahuja2019language2pose} & \et{0.246}{.002} & \et{0.387}{.002} & \et{0.486}{.002} & \et{11.02}{.046} & \et{5.296}{.008} & \et{7.676}{.058} & -  \\
        
        Text2Gesture~\cite{bhattacharya2021text2gestures} & \et{0.165}{.001} & \et{0.267}{.002} & \et{0.345}{.002} & \et{5.012}{.030} & \et{6.030}{.008} & \et{6.409}{.071} & -  \\
        
        Hier~\cite{ghosh2021synthesis} & \et{0.301}{.002} & \et{0.425}{.002} & \et{0.552}{.004} & \et{6.532}{.024} & \et{5.012}{.018} & \et{8.332}{.042} & -  \\

        MoCoGAN~\cite{tulyakov2018mocogan} & \et{0.037}{.000} & \et{0.072}{.001} & \et{0.106}{.001} & \et{94.41}{.021} & \et{9.643}{.006} & \et{0.462}{.008} & \et{0.019}{.000}  \\

        Dance2Music~\cite{lee2019dancing} & \et{0.033}{.000} & \et{0.065}{.001} & \et{0.097}{.001} & \et{66.98}{.016} & \et{8.116}{.006} & \et{0.725}{.011} & \et{0.043}{.001}  \\

        TEMOS$^\S$~\cite{petrovich22temos} & \et{0.424}{.002} & \et{0.612}{.002} & \et{0.722}{.002} & \et{3.734}{.028} & \et{3.703}{.008} & \et{8.973}{.071} & \et{0.368}{.018} \\
        
        TM2T~\cite{chuan2022tm2t} & \et{0.424}{.003} & \et{0.618}{.003} & \et{0.729}{.002} & \et{1.501}{.017} & \et{3.467}{.011} & \et{8.589}{.076} & \et{2.424}{.093}  \\
        
        Guo \textit{et al.}~\cite{guo2022generating} & \et{0.455}{.003} & \et{0.636}{.003} & \et{0.736}{.002} & \et{1.087}{.021} & \et{3.347}{.008} & \et{9.175}{.083} & \et{2.219}{.074}  \\ 

        MLD$^\S$~\cite{chen2022mld} & \et{0.481}{.003} & \et{0.673}{.003} & \et{0.772}{.002} & \et{0.473}{.013} & \et{3.196}{.010} & \et{9.724}{.082} & \et{2.413}{.079} \\

        MDM$^\S$~\cite{tevet2022MDM} & - & - & \et{0.611}{.007} & \et{0.544}{.044} & \et{5.566}{.027} & \et{9.559}{.086} & \etbb{2.799}{.072}  \\ 
        
        MotionDiffuse$^\S$~\cite{zhang2022motiondiffuse} & \etbb{0.491}{.001} & \etr{0.681}{.001} & \etr{0.782}{.001} & \et{0.630}{.001} & \etr{3.113}{.001} & \et{9.410}{.049} & \et{1.553}{.042}  \\ 
        
    \midrule
        Our GPT \small{($\tau = 0$)} & \et{0.417}{.003} & \et{0.589}{.002} & \et{0.685}{.003} & \etbb{0.140}{.006} & \et{3.730}{.009} & \etr{9.844}{.095} & \etr{3.285}{.070} \\
        Our GPT \small{($\tau = 0.5$)} & \etbb{0.491}{.003} & \etbb{0.680}{.003} & \etbb{0.775}{.002} & \etr{0.116}{.004} & \etbb{3.118}{.011} & \etbb{9.761}{.081} &  \et{1.856}{.011} \\
        Our GPT \small{($\tau \in \mathcal{U}[0, 1]$)} & \etr{0.492}{.003} & \et{0.679}{.002} & \etbb{0.775}{.002} & \et{0.141}{.005} & \et{3.121}{.009} & \et{9.722}{.082} &  \et{1.831}{.048} \\
    \bottomrule
    \end{tabular}
    }
    \vspace{-1mm}
    \caption{\textbf{Comparison with the state-of-the-art methods on HumanML3D~\cite{guo2022generating} test set.} We compute standard metrics following Guo \textit{et al.}~\cite{guo2022generating}. For each metric, we repeat the evaluation 20 times and report the average with 95\% confidence interval. \textcolor{red}{Red} and \textcolor{blue}{Blue} indicate the best and the second best result. $^\S$ reports results using ground-truth motion length.}
    \label{tab1}

\end{table*}

\begin{table*}[t]
    \centering
    \scalebox{0.955}{

    \begin{tabular}{l c c c c c c c}
    \toprule
    \multirow{2}{*}{Methods}  & \multicolumn{3}{c}{R-Precision $\uparrow$} & \multirow{2}{*}{FID $\downarrow$} & \multirow{2}{*}{MM-Dist $\downarrow$} & \multirow{2}{*}{Diversity $\uparrow$} & \multirow{2}{*}{MModality $\uparrow$}\\

    \cline{2-4}
    ~ & Top-1 & Top-2 & Top-3 \\
    
    \midrule
 \textbf{Real motion} & \et{0.424}{.005} & \et{0.649}{.006} & \et{0.779}{.006} & \et{0.031}{.004} & \et{2.788}{.012} & \et{11.08}{.097} & -  \\
 
Our VQ-VAE (Recons.) & \et{0.399}{.005} & \et{0.614}{.005} & \et{0.740}{.006} & \et{0.472}{.011} & \et{2.986}{.027} & \et{10.994}{.120} & -  \\
    \midrule
        Seq2Seq~\cite{lin2018generating} & \et{0.103}{.003} & \et{0.178}{.005} & \et{0.241}{.006} & \et{24.86}{.348} & \et{7.960}{.031} & \et{6.744}{.106}  & -  \\

        Language2Pose~\cite{ahuja2019language2pose} & \et{0.221}{.005} & \et{0.373}{.004} & \et{0.483}{.005} & \et{6.545}{.072} & \et{5.147}{.030} & \et{9.073}{.100} & -  \\

        Text2Gesture~\cite{bhattacharya2021text2gestures} & \et{0.156}{.004} & \et{0.255}{.004} & \et{0.338}{.005} & \et{12.12}{.183} & \et{6.964}{.029} & \et{9.334}{.079} & -  \\
        
        Hier~\cite{ghosh2021synthesis} & \et{0.255}{.006} & \et{0.432}{.007} & \et{0.531}{.007} & \et{5.203}{.107} & \et{4.986}{.027} & \et{9.563}{.072} & -  \\

         MoCoGAN~\cite{tulyakov2018mocogan} & \et{0.022}{.002} & \et{0.042}{.003} & \et{0.063}{.003} & \et{82.69}{.242} & \et{10.47}{.012} & \et{3.091}{.043} & \et{0.250}{.009}  \\

        Dance2Music~\cite{lee2019dancing} & \et{0.031}{.002} & \et{0.058}{.002} & \et{0.086}{.003} & \et{115.4}{.240} & \et{10.40}{.016} & \et{0.241}{.004} & \et{0.062}{.002}  \\

        TEMOS$^\S$~\cite{petrovich22temos,chen2022mld} & \et{0.353}{.002} & \et{0.561}{.002} & \et{0.687}{.002} & \et{3.717}{.028} & \et{3.417}{.008} & \et{10.84}{.071} & \et{0.532}{.018} \\
        
        TM2T~\cite{chuan2022tm2t} & \et{0.280}{.006} & \et{0.463}{.007} & \et{0.587}{.005} & \et{3.599}{.051} & \et{4.591}{.019} & \et{9.473}{.100} & \etr{3.292}{.034}  \\

        Guo \textit{et al.}~\cite{guo2022generating} & \et{0.361}{.006} & \et{0.559}{.007} & \et{0.681}{.007} & \et{3.022}{.107} & \et{3.488}{.028} & \et{10.72}{.145} & \et{2.052}{.107}  \\

        MLD$^\S$~\cite{chen2022mld} & \et{0.390}{.008} & \et{0.609}{.008} & \et{0.734}{.007} & \et{0.404}{.027} & \et{3.204}{.027} & \et{10.80}{.117} & \et{2.192}{.071} \\

        MDM$^\S$~\cite{tevet2022MDM} & - & - & \et{0.396}{.004} & \etr{0.497}{.021} & \et{9.191}{.022} & \et{10.847}{.109} & \et{1.907}{.214}  \\ 
        
        MotionDiffuse$^\S$~\cite{zhang2022motiondiffuse} & \etr{0.417}{.004} & \etbb{0.621}{.004} & \etbb{0.739}{.004} & \et{1.954}{.062} & \etr{2.958}{.005} & \etbb{11.10}{.143} & \et{0.730}{.013}  \\ 
    \midrule
        Our GPT ($\tau = 0$) & \et{0.392}{.007} & \et{0.600}{.007} & \et{0.716}{.006} & \et{0.737}{.049} & \et{3.237}{.027} & \etr{11.198}{.086} & \etbb{2.309}{.055}\\
        Our GPT ($\tau = 0.5$) & \et{0.402}{.006} & \et{0.619}{.005} & \et{0.737}{.006} & \et{0.717}{.041} & \et{3.053}{.026} & \et{10.862}{.094} & \et{1.912}{.036}\\
        Our GPT ($\tau \in \mathcal{U}[0, 1]$) & \etbb{0.416}{.006}  & \etr{0.627}{.006} & \etr{0.745}{.006} & \etbb{0.514}{.029} & \etbb{3.007}{.023} & \et{10.921}{.108} & \et{1.570}{.039} \\
    \bottomrule
    \end{tabular}
    }
    \vspace{-1mm}
    \caption{\textbf{Comparison with the state-of-the-art methods on KIT-ML~\cite{plappert2016kit} test set.} We compute standard metrics following Guo \textit{et al.}~\cite{guo2022generating}. For each metric, we repeat the evaluation 20 times and report the average with 95\% confidence interval. \textcolor{red}{Red} and \textcolor{blue}{Blue} indicate the best and the second best result.$^\S$ reports results using ground-truth motion length. }
    \label{tab2}
\end{table*}

\section{Experiment}
\label{sec:exp}
In this section, we present our experimental results. In Section~\ref{sec:data}, we introduce standard datasets as well as evaluation metrics. We compare our results to competitive approaches in Section~\ref{sec:compare}. Finally, we provide analysis and discussion in Section~\ref{sec:discuss}.

\subsection{Datasets and evaluation metric}
\label{sec:data}
We conduct experiments on two standard datasets for text-driven motion generations: HumanML3D~\cite{guo2022generating} and KIT Motion-Language (KIT-ML)~\cite{plappert2016kit}. Both datasets are commonly used in the community. We follow the evaluation protocol proposed in~\cite{guo2022generating}.

\noindent\textbf{KIT Motion-Language (KIT-ML).} KIT-ML~\cite{plappert2016kit} contains 3,911 human motion sequences and 6,278 textual annotations. The total vocabulary size, that is the number of unique words disregarding capitalization and punctuation, is 1,623. Motion sequences are selected from KIT~\cite{mandery2015kit} and CMU~\cite{cmu} datasets but downsampled into 12.5 frame-per-second (FPS). Each motion sequence is described by from 1 to 4 sentences. The average length of descriptions is approximately 8. Following~\cite{guo2022generating,chuan2022tm2t}, the dataset is split into training, validation, and test sets with proportions of 80\%, 5\%, and 15\%, respectively. We select the model that achieves the best FID on the validation set and report its performance on the test set.

\noindent\textbf{HumanML3D.} HumanML3D~\cite{guo2022generating} is currently the largest 3D human motion dataset with textual descriptions. The dataset contains 14,616 human motions and 44,970 text descriptions. The entire textual descriptions are composed of  5,371 distinct words. The motion sequences are originally from AMASS~\cite{mahmood2019amass} and HumanAct12~\cite{guo2020action2motion} but with specific pre-processing: motion is scaled to 20 FPS; those that are longer than 10 seconds are randomly cropped to 10-second ones; they are then re-targeted to a default human skeletal template and properly rotated to face Z+ direction initially. Each motion is paired with at least 3 precise textual descriptions. The average length of descriptions is approximately 12. According to~\cite{guo2022generating}, the dataset is split into training, validation, and test sets with proportions of 80\%, 5\%, and 15\%, respectively. 
We select the best FID model on the validation set and report its performance on the test set.

\paragraph{Implementation details.} 
For Motion VQ-VAE, the codebook size is set to $512 \times 512$. The downsampling rate $l$ is 4. 
We provide an ablation on the number of codes in Section~\ref{sec:supp_dilation_code} of the appendix. For both HumanML3D~\cite{guo2022generating} and KIT-ML~\cite{plappert2016kit} datasets, the motion sequences are cropped to $T=64$ for training. We use AdamW~\cite{loshchilov2018decoupled} optimizer with $[\beta_1, \beta_2] = [0.9, 0.99]$, batch size of 256, and exponential moving constant $\lambda=0.99$. We train the first 200K iterations with a learning rate of 2e-4, and 100K with a learning rate of 1e-5. $\beta$ and $\alpha$ in $\mathcal{L}_{\text{vq}}$ and $\mathcal{L}_{\text{re}}$ are set to 0.02 and 0.5, respectively. Following~\cite{guo2022generating}, the dataset KIT-ML and HumanML3D are extracted into motion features with dimensions 251 and 263 respectively, which correspond to local joints position, velocity, and rotations in root space as well as global translation and rotations. These features are computed from 21 and 22 joints of SMPL~\cite{loper2015smpl}. More details about the motion representations are provided in Section~\ref{sec:supp_metrics} of the appendix.

For the T2M-GPT, we employ 18 transformer~\cite{vaswani2017attention} layers with a dimension of 1,024 and 16 heads. 
The ablation for different scales of the transformer is provided in Section~\ref{sec:supp_trans_arch} of the appendix. 
Following Guo \textit{et al}.~\cite{guo2022generating}, the maximum length of Motion is 196 on both datasets, and the minimum lengths are 40 and 24 for HumanML3D~\cite{guo2022generating} and KIT-ML~\cite{plappert2016kit} respectively. The maximum length of the code index sequence is $T'=50$. We train an extra $\mathit{End}$ token as a signal to stop index generation. The transformer is optimized using AdamW~\cite{loshchilov2018decoupled} with $[\beta_1, \beta_2] = [0.5, 0.99]$ and batch size 128. The initialized learning rate is set to 1e-4 for 150K iterations and decayed to 5e-6 for another 150K iterations. Training Motion VQ-VAE and T2M-GPT take about 14 hours and 78 hours respectively on a single Tesla V100-32G GPU. 

\begin{figure*}[tp]
    \centering
    \includegraphics[width=0.92\textwidth]{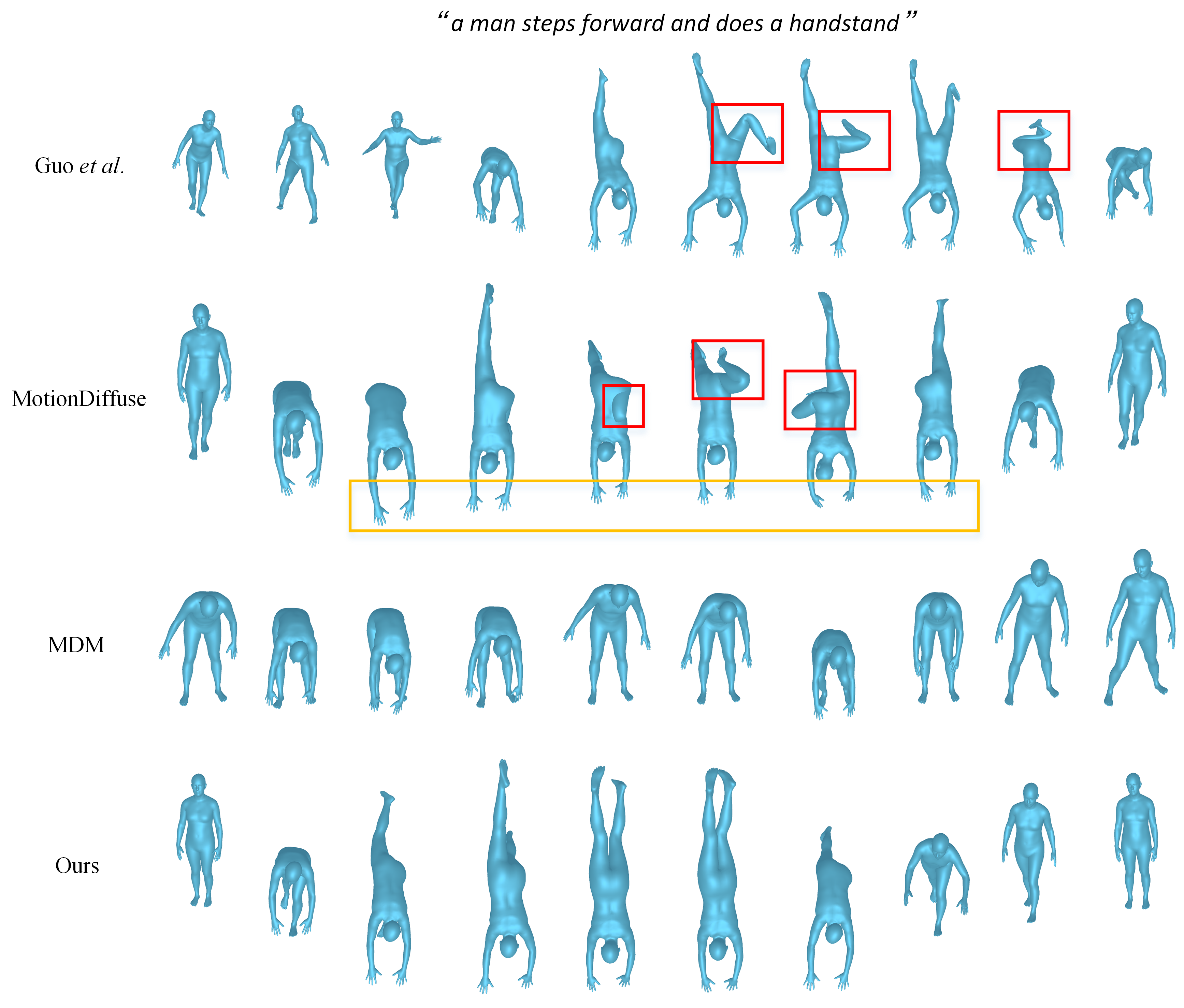}
    \caption{\textbf{Visual results on HumanML3D~\cite{guo2022generating} dataset.} We compare our generation with Guo \textit{et al.}~\cite{guo2022generating}, MotionDiffuse~\cite{zhang2022motiondiffuse}, and MDM~\cite{tevet2022MDM}. Distorted motions (\textcolor{red}{red}) and sliding (\textcolor{yellow}{yellow}) are highlighted. More visual results can be found on the \href{https://mael-zys.github.io/T2M-GPT/}{project page}.}
    \label{fig:visual}
\end{figure*}

\paragraph{Evaluation metric.} Following~\cite{guo2022generating}, global representations of motion and text descriptions are first extracted with the pre-trained network in~\cite{guo2022generating}, and then measured by the following five metrics:
\begin{itemize}
  \item \textit{R-Precision.} Given one motion sequence and 32 text descriptions (1 ground-truth and 31 randomly selected mismatched descriptions), we rank the Euclidean distances between the motion and text embeddings. Top-1, Top-2, and Top-3 accuracy of motion-to-text retrieval are reported.
  \item \textit{Frechet Inception Distance (FID).} We calculate the distribution distance between the generated and real motion using FID~\cite{heusel2017gans} on the extracted motion features.
  \item \textit{Multimodal Distance (MM-Dist).} The average Euclidean distances between each text feature and the generated motion feature from this text.
  \item \textit{Diversity.} From a set of motions, we randomly sample 300 pairs of motion. We extract motion features and compute the average Euclidean distances of the pairs to measure motion diversity in the set.
  \item \textit{Multimodality (MModality).} For one text description, we generate 20 motion sequences forming 10 pairs of motion. We extract motion features and compute the average Euclidean distances of the pairs. We finally report the average over all the text descriptions.
\end{itemize}

Note that more details about the evaluation metrics are provided in Section~\ref{sec:supp_metrics} of the appendix.

\subsection{Comparison to state-of-the-art approaches}
\label{sec:compare}

\noindent\textbf{Quantitative results.} 
We show the comparison results in Table~\ref{tab1} and Table~\ref{tab2} on HumanML3D~\cite{guo2022generating} test set and KIT-ML~\cite{plappert2016kit} test set. On both datasets, our reconstruction with VQ-VAE reaches close performances to real motion, which suggests high-quality discrete representations learned by our VQ-VAE. For the generation, our approach achieves comparable performance on text-motion consistency (R-Precision and MM-Dist) compared to the state-of-the-art method MotionDiffuse~\cite{zhang2022motiondiffuse}, while significantly outperforms MotionDiffuse with FID metric. KIT-ML~\cite{plappert2016kit} and HumanML3D~\cite{guo2022generating} are in different scales, which demonstrates the robustness of the proposed approach. Manually corrupting sequences during the training of GPT brings consistent improvement ($\tau = 0.5$ v.s.~$\tau = 0$). A more detailed analysis is provided in Section~\ref{sec:supp_rep_ratio} of the appendix. Unlike Guo \textit{et al.}~\cite{guo2022generating} involving an extra module to predict motion length, we implicitly learn the motion length through an additional $\mathit{End}$ token, which is more straightforward and shown to be more effective. Note that MDM~\cite{tevet2022MDM} and MotionDiffuse~\cite{zhang2022motiondiffuse} evaluate their models with the ground-truth motion length, which is not practical for real applications. 

\paragraph{Qualitative comparison.}
Figure~\ref{fig:visual} shows visual results on HumanML3D~\cite{guo2022generating}. We compare our generations with the current state-of-the-art models: Guo \textit{et al.}~\cite{guo2022generating}, MDM~\cite{tevet2022MDM} and MotionDiffuse~\cite{zhang2022motiondiffuse}. From the example in Figure~\ref{fig:visual}, one can figure out that our model generates human motion with better quality than the others, and we highlight in red for unrealistic motion generated by Guo \textit{et al.}~\cite{guo2022generating} and MotionDiffuse~\cite{zhang2022motiondiffuse}. Moreover, the generated motion of MDM~\cite{tevet2022MDM} is not related to the text description. Note that more visual results and the failure case are provided on the \href{https://mael-zys.github.io/T2M-GPT/}{project page}.

\begin{table}[t]
\centering
    \resizebox{0.48\textwidth}{!}{
    \begin{tabular}{cc|cc|cc}
        \toprule
        \multicolumn{2}{c|}{Quantizer} & \multicolumn{2}{c|}{Reconstruction} &  \multicolumn{2}{c}{Generation} \\
           Code Reset & EMA & FID $\downarrow$ & Top-1 $\uparrow$ & FID $\downarrow$ & Top-1 $\uparrow$ \\ 
           \midrule
            &  & \et{0.492}{.004} & \et{0.436}{.003} & \et{42.797}{.156} & \et{0.048}{.001} \\
            & \cmark & \et{0.097}{.001} & \et{0.499}{.002} & \et{0.176}{.008} & \et{0.490}{.002} \\
           \cmark &  & \et{0.102}{.001} & \et{0.494}{.003} & \et{0.248}{.009} & \et{0.461}{.002} \\
           \cmark & \cmark & \etb{0.070}{.001} & \etb{0.501}{.002} & \etb{0.116}{.004} & \etb{0.491}{.003}  \\
       \bottomrule
    \end{tabular}
    }
    \caption{\textbf{Analysis of VQ-VAE quantizers on HumanML3D~\cite{guo2022generating} test set.} For all the quantizers, we set $\tau=0.5$ and use the same architectures (VQ-VAE and GPT) described in Section~\ref{sec:data}. We report FID and Top-1 for both reconstruction and generation. For each metric, we repeat the evaluation 20 times and report the average with 95\% confidence interval.}
    \label{tab:vq_analysis}
\end{table}  

\begin{figure}[t]
 \centering
  \begin{subfigure}[b]{0.49\columnwidth}
    \includegraphics[width=\linewidth]{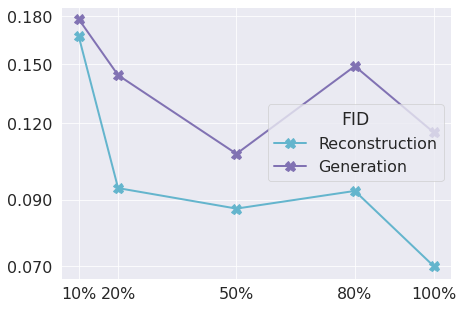}
    \caption{FID}
    \label{fig:fid}
  \end{subfigure}
  \hfill 
  \begin{subfigure}[b]{0.49\columnwidth}
    \includegraphics[width=\linewidth]{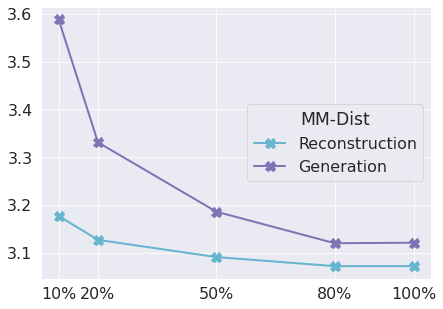}
    \caption{MM-Dist}
    \label{fig:mmdist}
  \end{subfigure}
  \hfill 
  \begin{subfigure}[b]{0.49\columnwidth}
    \includegraphics[width=\linewidth]{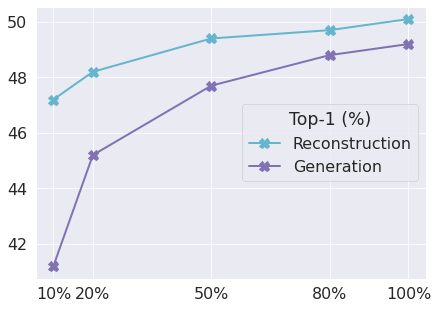}
    \caption{Top-1 accuracy}
    \label{fig:top1}
  \end{subfigure}
  \hfill 
  \begin{subfigure}[b]{0.49\columnwidth}
    \includegraphics[width=\linewidth]{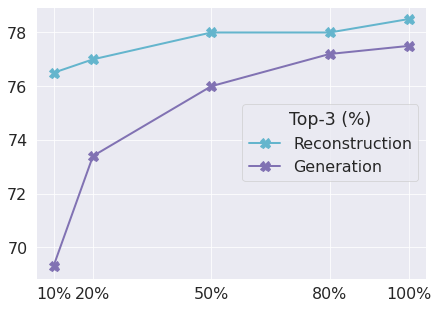}
    \caption{Top-3 accuracy}
    \label{fig:top3}
  \end{subfigure}
  \caption{\textbf{Impact of dataset size on HumanML3D~\cite{guo2022generating}.} We train our motion VQ-VAE (\textit{Reconstruction}) and  T2M-GPT (\textit{Generation}) on the subsets of HumanML3D~\cite{guo2022generating} composed of 10\%, 20\%, 50\%, 80\%, and 100\% training set respectively. All the models are evaluated on the entire test set. We report FID, MM-Dist, Top-1, and Top-3 accuracy for all the models. Results suggest that our model might benefit from more training data.}
  \label{fig:dataset}
\end{figure}

\subsection{Discussion}
\label{sec:discuss}
\paragraph{Quantization strategies.} We first investigate the impact of different quantization strategies presented in Section~\ref{sec:vqvae}. The results are illustrated in Table~\ref{tab:vq_analysis} for both reconstruction and generation. We notice that naive VQ-VAE training is not able to reconstruct nor generate high-quality motion. However, training with \textit{EMA} or \textit{Code Reset} can importantly boost the performances for both reconstruction and generation. 

\paragraph{Impact of dataset size.} We further analyze the impact of dataset size. To understand whether the largest dataset HumanML3D~\cite{guo2022generating} contains enough data for motion generation, we train our motion VQ-VAE and T2M-GPT on different subsets of the training data, which consists of 10\%, 20\%, 50\%, 80\% and 100\% of the training data respectively. The trained models are evaluated on the entire test set. The results are illustrated in Figure~\ref{fig:dataset}. We evaluate reconstruction for our motion VQ-VAE and generation for our T2M-GPT using four metrics: FID, MM-Dist, Top-1, and Top-3 accuracies. Several insights can be figured out: \textit{i)} metric for motion quality (FID) and metric for motion-text consistency (MM-Dist, Top-1, and Top-3) should be considered at the same time. With only 10\% data, the motion might be of good quality, however, the model is not able to generate a correct motion that corresponds to the text description; \textit{ii)} the performances become better with more training data. This trend suggests that additional training data could bring non-negligible improvement to both reconstruction and generation.

\section{Conclusion}
\label{sec:con}

In this work, we investigated a classic framework based on VQ-VAE and GPT to synthesize human motion from textual descriptions. Our method achieved comparable or even better performances than concurrent diffusion-based approaches, 
suggesting that this classic framework remains a very competitive approach for motion generation. We explored in detail the effect of various quantization strategies on motion reconstruction and generation. Moreover, we provided an analysis of the dataset size. Our finding suggests that a larger dataset could still bring additional improvement to our approach.
\paragraph{Acknowledgement} We thank Mathis Petrovich, Yuming Du, Yingyi Chen, Dexiong Chen, and Xuelin Chen for inspiring discussions and valuable feedback. This paper is supported by NSF of China (No. 62176155) and Jilin Province (20200201037JC), \textit{etc}. More funding information is provided in Section~\ref{sec:supp_fund} of the appendix.

{\small
\bibliographystyle{ieee_fullname}
\bibliography{egbib}
}

\clearpage
\appendix
\vspace*{1em}{\centering\large\bf%
Appendix
\vspace*{1.5em}}

In this appendix, we present:
\begin{itemize}
    \item Section~\ref{sec:supp_trans_arch}: ablation study of T2M-GPT architecture.
    \item Section~\ref{sec:supp_vq_loss}: ablation study of the reconstruction loss ($\mathcal{L}_{re}$ in Equation [3]) for motion VQ-VAE.
    \item Section~\ref{sec:supp_rep_ratio}: ablation study of $\tau$ for the corruption strategy in T2M-GPT training.
    \item Section~\ref{sec:supp_dilation_code}: ablation study of the number of codes in VQ-VAE.
    \item Section~\ref{sec:supp_metrics}: more details on the evaluation metrics and the motion representations.
    \item Section~\ref{sec:supp_vq_arch}: the detail of the Motion VQ-VAE architecture.
    \item Section~\ref{sec:supp_limitation}: limitations of our proposed approach.
    \item Section~\ref{sec:supp_fund}: more funding information.
\end{itemize}

\section{Ablation study of T2M-GPT architecture}
\label{sec:supp_trans_arch}

\begin{table*}[t]
    \centering\setlength{\tabcolsep}{12pt}
    \resizebox{0.8\textwidth}{!}{
    \begin{tabular}{cccccc}
        \toprule
        Num. layers & Num. dim & Num. heads & FID $\downarrow$ & Top-1 $\uparrow$ & Training time (hours). \\
        \midrule
        4 & 512 & 8 & \et{0.469}{.014} & \et{0.469}{.002} & 17 \\  
        8 & 512 & 8 & \et{0.339}{.010} & \et{0.481}{.002} & 23 \\
        8 & 768 & 8 & \et{0.338}{.009} & \et{0.490}{.003} & 30 \\
        8 & 768 & 12 & \et{0.296}{.009} & \et{0.484}{.002} & 31 \\
        12 & 768 & 12 & \et{0.273}{.007} & \et{0.487}{.002} & 40 \\
        12 & 1024 & 16 & \et{0.149}{.007} & \et{0.489}{.002} & 55 \\
        16 & 768 & 12 & \et{0.145}{.006} & \et{0.486}{.003} & 47 \\
        16 & 1024 & 16 & \et{0.143}{.007} & \et{0.490}{.004}  & 59 \\
        18 & 768 & 12 & \etb{0.130}{.006} & \et{0.483}{.003} & 51 \\
        18 & 1024 & 16 & \et{0.141}{.005} & \etb{0.492}{.003} & 78 \\
        \bottomrule
    \end{tabular}
    }
    \caption{\textbf{Ablation study of T2M-GPT architecture on HumanML3D~\cite{guo2022generating} test set.} For all the architectures, we use the same motion VQ-VAE. The T2M-GPT is trained with $\tau \in \mathcal{U}[0, 1]$. The training time is evaluated on a single Tesla V100-32G GPU.}
    \label{tab:supp_trans_arch}
\end{table*}

In this section, we present results with different transformer architectures for T2M-GPT. The results are provided in Table~\ref{tab:supp_trans_arch}. We notice that better performance can be obtained with a larger architecture. We finally leverage an 18-layer transformer with 16 heads and 1,024 dimensions. 

\section{Impact of the reconstruction loss in motion VQ-VAE}
\label{sec:supp_vq_loss}

\begin{table}[h]
    \centering\setlength{\tabcolsep}{10pt}
    \resizebox{0.4\textwidth}{!}{
    \begin{tabular}{lccc}
        \toprule
        \multirow{2}{*}{$\mathcal{L}_{cons}$} & \multirow{2}{*}{$\alpha$} &
        \multicolumn{2}{c}{Reconstruction} \\ \cmidrule{3-4}
         & & FID $\downarrow$ & Top-1 (\%)  \\ \midrule
        L1 & 0 & \et{0.095}{.001} & \et{0.493}{.002}  \\
        L1 & 0.5 & \et{0.144}{.001} & \et{0.495}{.003}  \\
        L1 & 1 & \et{0.160}{.001} & \et{0.496}{.003}  \\
        \midrule
        L1Smooth & 0 & \et{0.112}{.001} & \et{0.496}{.003}  \\
        L1Smooth & 0.5 & \etb{0.070}{.001} & \etb{0.501}{.002}  \\
        L1Smooth & 1 & \et{0.128}{.001} & \et{0.499}{.003}  \\
        \midrule
        L2 & 0 & \et{0.321}{.002} & \et{0.478}{.003}  \\
        L2 & 0.5 & \et{0.292}{.002} & \et{0.483}{.002}  \\
        L2 & 1 & \et{0.213}{.002} & \et{0.490}{.003} \\
        
         \bottomrule
    \end{tabular}
    }
    \caption{\textbf{Ablation of losses for VQ-VAE on HumanML3D~\cite{guo2022generating} test set.} We report FID and Top1 metric for the models trained 300K iterations.}
\label{tab:supp_loss}
\end{table}

In this section, we study the effect of the reconstruction loss ($\mathcal{L}_{re}$ in Equation [3]) and the hyper-parameter $\alpha$ (Equation [3]). The results are presented in Table~\ref{tab:supp_loss}. We find that L1 Smooth achieves the best performance on reconstruction, and the performance of L1 loss is close to L1 Smooth loss. For the hyper-parameter $\alpha$, we find that $\alpha = 0.5$ leads to the best performance.

\section{Impact of $\tau$ for the corruption strategy in T2M-GPT training}
\label{sec:supp_rep_ratio}

\begin{table}[h]
    \centering\setlength{\tabcolsep}{12pt}
    \resizebox{0.5\textwidth}{!}{
    \begin{tabular}{cccc}
    \toprule
        $\tau$  & FID $\downarrow$ & Top-1 $\uparrow$  & MM-Dist $\downarrow$ \\
        \midrule
        $0.0$ & \et{0.140}{.006} & \et{0.417}{.003} & \et{3.730}{.009}\\
        $0.1$ & \et{0.131}{.005} & \et{0.453}{.002} & \et{3.357}{.007}\\
        $0.3$ & \et{0.147}{.006} & \et{0.485}{.002} & \et{3.157}{.007}\\
        $0.5$ & \etb{0.116}{.004} & \et{0.491}{.003} & \etb{3.118}{.011}\\ 
        $0.7$ & \et{0.155}{.006} & \et{0.480}{.004} & \et{3.183}{.011}\\
        $\mathcal{U}[0, 1]$ & \et{0.141}{.005} & \etb{0.492}{.003} & \et{3.121}{.009} \\ 
        \bottomrule
    \end{tabular}
    }
    \caption{\textbf{Analysis of $\tau$ on HumanML3D~\cite{guo2022generating} test set.}}
    \label{tab:mask}
\end{table}

In this section, we study $\tau$, which is used for corrupting sequences during the training of T2M-GPT. The results are provided in Table~\ref{tab:mask}. We can see that the training with corrupted sequences $\tau = 0.5$ significantly improves over Top-1 accuracy and FID compared to $\tau = 0$. Compared to $\tau \in \mathcal{U}[0, 1]$, $\tau = 0.5$ is probably preferable for HumanML3D~\cite{guo2022generating}, as it achieves comparable Top-1 accuracy compared to $\tau \in \mathcal{U}[0, 1]$ but with much better FID. 

\section{Ablation study of the number of codes in VQ-VAE}
\label{sec:supp_dilation_code}

\begin{table}[t]
    \centering\setlength{\tabcolsep}{12pt}
    \resizebox{0.4\textwidth}{!}{
    \begin{tabular}{ccc}
    \toprule
        \multirow{2}{*}{Num. code} & 
        \multicolumn{2}{c}{Reconstruction}  \\ \cmidrule{2-3}
         &  FID $\downarrow$ & Top-1 (\%)  \\ \midrule
        256 & \et{0.145}{.001} & \et{0.497}{.002}  \\
        512 & \etb{0.070}{.001} & \etb{0.501}{.002}  \\
        1024 & \et{0.090}{.001} & \et{0.498}{.003}  \\ \bottomrule
    \end{tabular}
    }
    \caption{\textbf{Study on the number of code in codebook on HumanML3D~\cite{guo2022generating} test set.}}
    \label{tab:supp_codenum}
\end{table}

We investigate the number of codes in the codebook in Table~\ref{tab:supp_codenum}. We find that the performance of 512 codes is slightly better than 1,024 codes. The results show that 256 codes are not sufficient for reconstruction.

\section{More details on the evaluation metrics and the motion representations.}
\label{sec:supp_metrics}

\subsection{Evaluation metrics}

We detail the calculation of several evaluation metrics, which are proposed in ~\cite{guo2022generating}. We denote ground-truth motion features, generated motion features, and text features as $f_{gt}$, $f_{pred}$, and $f_{text}$. Note that these features are extracted with pretrained networks in~\cite{guo2022generating}.

\paragraph{FID.}
FID is widely used to evaluate the overall quality of the generation. We obtain FID by
\begin{equation}
\text{FID} = \lVert \mu_{gt} - \mu_{pred}\rVert^2 - \text{Tr}(\Sigma_{gt} + \Sigma_{pred} - 2(\Sigma_{gt}\Sigma_{pred})^{\frac{1}{2}})
\label{formula:fid}
\end{equation}
where $\mu_{gt}$ and $\mu_{pred}$ are mean of $f_{gt}$ and $f_{pred}$. $\Sigma$ is the covariance matrix and $\text{Tr}$ denotes the trace of a matrix.

\paragraph{MM-Dist.}
MM-Dist measures the distance between the text embedding and the generated motion feature. Given N randomly generated samples, the MM-Dist measures the feature-level distance between the motion and the text. Precisely, it computes the average Euclidean distances between each text feature and the
generated motion feature as follows:
\begin{equation}
\text{MM-Dist} = \frac{1}{N}\sum_{i=1}^{N}\lVert f_{pred,i} - f_{text,i}\rVert
\label{formula:mm-dis}
\end{equation}
where $f_{pred,i}$ and  $f_{text,i}$ are the features of the i-th text-motion pair. 

\paragraph{Diversity.} Diversity measures the variance of the whole motion sequences across the dataset. We randomly sample $S_{dis}$ pairs of motion and each pair of motion features is denoted by $f_{pred,i}$ and $f_{pred,i}'$. The diversity can be calculated by
\begin{equation}
\text{Diversity} = \frac{1}{S_{dis}}\sum_{i=1}^{S_{dis}}||f_{pred,i} - f_{pred,i}'||
\label{formula:diversity}
\end{equation}
In our experiments, we set $S_{dis}$ to 300 as \cite{guo2022generating}.

\paragraph{MModality.} MModality measures the diversity of human motion generated from the same text description. Precisely, for the i-th text description, we generate motion 30 times and then sample two subsets containing 10 motion. We denote features of the j-th pair of the i-th text description by ($f_{pred,i,j}$, $f_{pred,i,j}'$). The MModality is defined as follows:
\begin{equation}
\text{MModality} = \frac{1}{10N}\sum_{i=1}^{N}\sum_{j=1}^{10}\lVert f_{pred,i,j} - f_{pred,i,j}'\rVert
\label{formula:mmodality}
\end{equation}

\subsection{Motion representations}

We use the same motion representations as ~\cite{guo2022generating}. Each pose is represented by $(\dot{r}^a, \dot{r}^x, \dot{r}^z, r^y, j^p, j^v, j^r, c^f)$, where $\dot{r}^a \in \mathbb{R}$ is the global root angular velocity; $\dot{r}^x \in \mathbb{R}, \dot{r}^z \in \mathbb{R}$ are the global root velocity in the X-Z plan; $j^p \in \mathbb{R}^{3j}, j^v \in \mathbb{R}^{3j}, j^r \in \mathbb{R}^{6j}$ are the local pose positions, velocity and rotation with j the number of joints; $c^f\in \mathbb{R}^4$ is the foot contact features calculated by the heel and toe joint velocity.

\section{VQ-VAE Architecture}
\label{sec:supp_vq_arch}
We illustrate the detailed architecture of VQ-VAE in Table~\ref{tab:supp_vqarch}. The dimensions of the HumanML3D~\cite{guo2022generating} and KIT-ML~\cite{plappert2016kit} datasets feature are 263 and 259 respectively.

\begin{table}[t]
    \centering\setlength{\tabcolsep}{12pt}
    \resizebox{0.4\textwidth}{!}{
    \begin{tabular}{ccc}
    \toprule
        \multirow{2}{*}{Dilation rate} & 
        \multicolumn{2}{c}{Reconstruction} \\ \cmidrule{2-3}
         &  FID $\downarrow$ & Top-1 (\%)  \\ \midrule
        1, 1, 1 & \et{0.145}{.001} & \et{0.500}{.003}  \\
        4, 2, 1 & \et{0.138}{.001} & \etb{0.502}{.002}  \\
        9, 3, 1 & \etb{0.070}{.001} & \et{0.501}{.002}  \\ 
        16, 4, 1 & \et{57.016}{.084} & \et{0.032}{.001} \\\bottomrule
    \end{tabular}
    }
    \caption{\textbf{Ablation study of different dilation rate in VQ-VAE on HumanML3D~\cite{guo2022generating} test set.}}
    \label{tab:supp_dilation}
\end{table}

\paragraph{Dilation rate.}

We investigate the impact of different dilation rates of the convolution layers used in VQ-VAE, and the results are presented in Table~\ref{tab:supp_dilation} for reconstruction. We notice that setting the dilation rate as (9, 3, 1) gives the most effective and stable performance.

\section{Limitations}
\label{sec:supp_limitation}
Our approach has two limitations: \textit{i)} for excessively long texts, the generated motion might miss some details of the textual description. Note that this typical failure case exists for all competitive approaches. \textit{ii)} some generated motion sequences slightly jitter on the legs and hands movement, this can be seen from the visual results provided in the appendix. We think the problem comes from the VQ-VAE architecture, with a better-designed architecture, the problem might be alleviated. For a real application, the jittering problem could be addressed using a temporal smoothing filter as a post-processing step.

\section{Funding Support}
\label{sec:supp_fund}
This work is supported by:
\begin{itemize}
    \item Natural Science Foundation of China (No. 62176155).
    \item Natural Science Foundation of Jilin Province (20200201037JC).
    \item Provincial Science and Technology Innovation Special Fund Project of Jilin Province (20190302026GX)
    \item Shanghai Municipal Science and Technology Major Project (2021SHZDZX0102)
\end{itemize}

\begin{table*}[h]
    \centering\setlength{\tabcolsep}{12pt}
    \begin{tabular}{ll}
    \toprule
        Components & Architecture \\ \midrule
        VQ-VAE Encoder & (0): Conv1D($D_{in}$, 512, kernel\_size=(3,), stride=(1,), padding=(1,)) \\
        ~ & (1): ReLU() \\
        ~ & (2): 2 $\times$ Sequential( \\
        ~ &   ~~~~(0): Conv1D(512, 512, kernel\_size=(4,), stride=(2,), padding=(1,)) \\
        ~ &   ~~~~(1): Resnet1D( \\
        ~ &   ~~~~~~~~    (0): ResConv1DBlock( \\
        ~ &   ~~~~~~~~~~~~      (activation1): ReLU() \\
        ~ &   ~~~~~~~~~~~~      (conv1): Conv1D(512, 512, kernel\_size=(3,), stride=(1,), padding=(9,), dilation=(9,)) \\
        ~ &   ~~~~~~~~~~~~      (activation2): ReLU() \\
        ~ &   ~~~~~~~~~~~~      (conv2): Conv1D(512, 512, kernel\_size=(1,), stride=(1,))) \\
        ~ &   ~~~~~~~~    (1): ResConv1DBlock( \\
        ~ &   ~~~~~~~~~~~~      (activation1): ReLU() \\
        ~ &    ~~~~~~~~~~~~     (conv1): Conv1D(512, 512, kernel\_size=(3,), stride=(1,), padding=(3,), dilation=(3,)) \\
        ~ &   ~~~~~~~~~~~~      (activation2): ReLU() \\
        ~ &    ~~~~~~~~~~~~     (conv2): Conv1D(512, 512, kernel\_size=(1,), stride=(1,))) \\
        ~ &   ~~~~~~~~    (2): ResConv1DBlock( \\
        ~ &   ~~~~~~~~~~~~      (activation1): ReLU() \\
        ~ &   ~~~~~~~~~~~~      (conv1): Conv1D(512, 512, kernel\_size=(3,), stride=(1,), padding=(1,)) \\
        ~ &   ~~~~~~~~~~~~      (activation2): ReLU() \\
        ~ &   ~~~~~~~~~~~~      (conv2): Conv1D(512, 512, kernel\_size=(1,), stride=(1,))))) \\
        \midrule
        Codebook & nn.Parameter((512, 512), requires\_grad=False) \\
        \midrule
        VQ-VAE Decoder & (0): 2 $\times$ Sequential( \\
        ~ &   ~~~~(0): Conv1D(512, 512, kernel\_size=(3,), stride=(1,), padding=(1,)) \\
        ~ &   ~~~~(1): Resnet1D( \\
        ~ &   ~~~~~~~~    (0): ResConv1DBlock( \\
        ~ &   ~~~~~~~~~~~~      (activation1): ReLU() \\
        ~ &   ~~~~~~~~~~~~      (conv1): Conv1D(512, 512, kernel\_size=(3,), stride=(1,), padding=(9,), dilation=(9,)) \\
        ~ &   ~~~~~~~~~~~~      (activation2): ReLU() \\
        ~ &   ~~~~~~~~~~~~      (conv2): Conv1D(512, 512, kernel\_size=(1,), stride=(1,))) \\
        ~ &   ~~~~~~~~    (1): ResConv1DBlock( \\
        ~ &   ~~~~~~~~~~~~      (activation1): ReLU() \\
        ~ &    ~~~~~~~~~~~~     (conv1): Conv1D(512, 512, kernel\_size=(3,), stride=(1,), padding=(3,), dilation=(3,)) \\
        ~ &   ~~~~~~~~~~~~      (activation2): ReLU() \\
        ~ &    ~~~~~~~~~~~~     (conv2): Conv1D(512, 512, kernel\_size=(1,), stride=(1,))) \\
        ~ &   ~~~~~~~~    (2): ResConv1DBlock( \\
        ~ &   ~~~~~~~~~~~~      (activation1): ReLU() \\
        ~ &   ~~~~~~~~~~~~      (conv1): Conv1D(512, 512, kernel\_size=(3,), stride=(1,), padding=(1,)) \\
        ~ &   ~~~~~~~~~~~~      (activation2): ReLU() \\
        ~ &   ~~~~~~~~~~~~      (conv2): Conv1D(512, 512, kernel\_size=(1,), stride=(1,))))) \\
        ~ &   ~~~~(2): Upsample(scale\_factor=2.0, mode=nearest) \\
        ~ &   ~~~~(3): Conv1D(512, 512, kernel\_size=(3,), stride=(1,), padding=(1,)) \\
        ~ & (1): ReLU() \\
        ~ & (2): Conv1D(512, $D_{in}$, kernel\_size=(3,), stride=(1,), padding=(1,))
 \\
         \bottomrule
    \end{tabular}
    \caption{\textbf{Architecture of our Motion VQ-VAE.}}
    \label{tab:supp_vqarch}
\end{table*}

\end{document}